\documentclass[letterpaper]{article} 
\usepackage{aaai2027}  
\usepackage[hyphens]{url}  
\usepackage{graphicx} 
\usepackage{natbib}  
\usepackage{caption} 
\usepackage{algorithm}
\usepackage{algorithmic}

\usepackage{newfloat}
\usepackage{listings}
\DeclareCaptionStyle{ruled}{labelfont=normalfont,labelsep=colon,strut=off} 
\floatstyle{ruled}
\newfloat{listing}{tb}{lst}{}
\floatname{listing}{Listing}

\usepackage{booktabs}
\usepackage{booktabs}
\usepackage{cite}
\usepackage{amsmath,amssymb,amsfonts}
\usepackage{algorithmic}
\usepackage{graphicx}
\usepackage{subcaption}
\usepackage{textcomp}
\usepackage{xcolor}
\usepackage{booktabs}
\usepackage{pgfplots}
\pgfplotsset{compat=1.18}
\usepackage{threeparttable}
\usepackage{tabularx}

\title{ZVeC: A Zero-Shot Framework for Instance-Level Vehicle Extraction and Generative Point Cloud Completion}
\author {
    Daisy Li\textsuperscript{\rm 1},
    Kyle Gao\textsuperscript{\rm 1},
    Quanyun Wu\textsuperscript{\rm 1},
    Boris Jutzi\textsuperscript{\rm 2},
    John S. Zelek\textsuperscript{\rm 1}\corresponding,
    Jonathan Li\textsuperscript{\rm 1}\corresponding,
}
\affiliations {
    \textsuperscript{\rm 1}University of Waterloo\\
    \textsuperscript{\rm 2}Karlsruhe Institute of Technology\\
    c779li@uwaterloo.ca, y56gao@uwaterloo.ca, quanyun.wu@uwaterloo.ca,  boris.jutzi@kit.edu, jzelek@uwaterloo.ca, junli@uwaterloo.ca
}

\begin{document}

\maketitle

\begin{abstract}
LiDAR point clouds acquired in underground environments exhibit severe geometric incompleteness due to occlusions and limited sensor viewpoints, making reliable point cloud completion challenging without large supervised datasets. We propose \textbf{ZVeC}, a zero-shot, instance-driven framework that reformulates scene-level completion as compositional object-level reconstruction. By decomposing a scene into semantic object instances, ZVeC reduces reconstruction ambiguity in cluttered environments while eliminating the need for scenario-specific training. Each segmented vehicle is completed independently using a depth- and 3D Gaussian-conditioned diffusion model that exploits generalized geometric priors before the reconstructed instances are recomposed into the original scene. To evaluate our approach, we construct a real-world dense LiDAR benchmark of underground parking environments. Experimental results demonstrate consistent improvements over representative scene-level baselines in both quantitative metrics and visual quality. The completed point cloud differs substantially from the measured input (average KL divergence $\sim$ 2.1), yet reducing the input to only 1\% of the original LiDAR measurements changes the completed reconstruction only marginally (KL divergence $< 0.50$). This demonstrates that ZVeC produces geometrically consistent completions even under extreme input sparsity.
\end{abstract}

\section{Introduction}

Reliable understanding of LiDAR point clouds is fundamental to applications including autonomous driving, robotic navigation, and 3D digital twin construction~\cite{b1,b2}. However, point clouds acquired from a single LiDAR pass often suffer from severe geometric incompleteness due to limited viewpoints, weak returns, and heavy occlusions~\cite{b3,b4,b5}. Since LiDAR cannot penetrate solid objects, large portions of object geometry remain unobserved, making point cloud completion a critical step for reliable scene understanding and reconstruction~\cite{b6,b7}.

Traditional point cloud completion methods rely on geometric assumptions such as symmetry, surface smoothness, or template matching, limiting their ability to recover heavily occluded structures. Recent learning-based methods generally follow either scene-level or instance-level paradigms~\cite{b8}. Scene-level approaches reconstruct entire environments but struggle in cluttered scenes where interactions among neighboring objects introduce significant ambiguity. Instance-level methods alleviate this issue by exploiting category-level shape priors. However, most existing approaches require supervised training on large paired datasets collected under specific acquisition settings, reducing their robustness to domain shifts encountered in real-world deployments.

\begin{figure}
    \centering
    \includegraphics[width=1\linewidth]{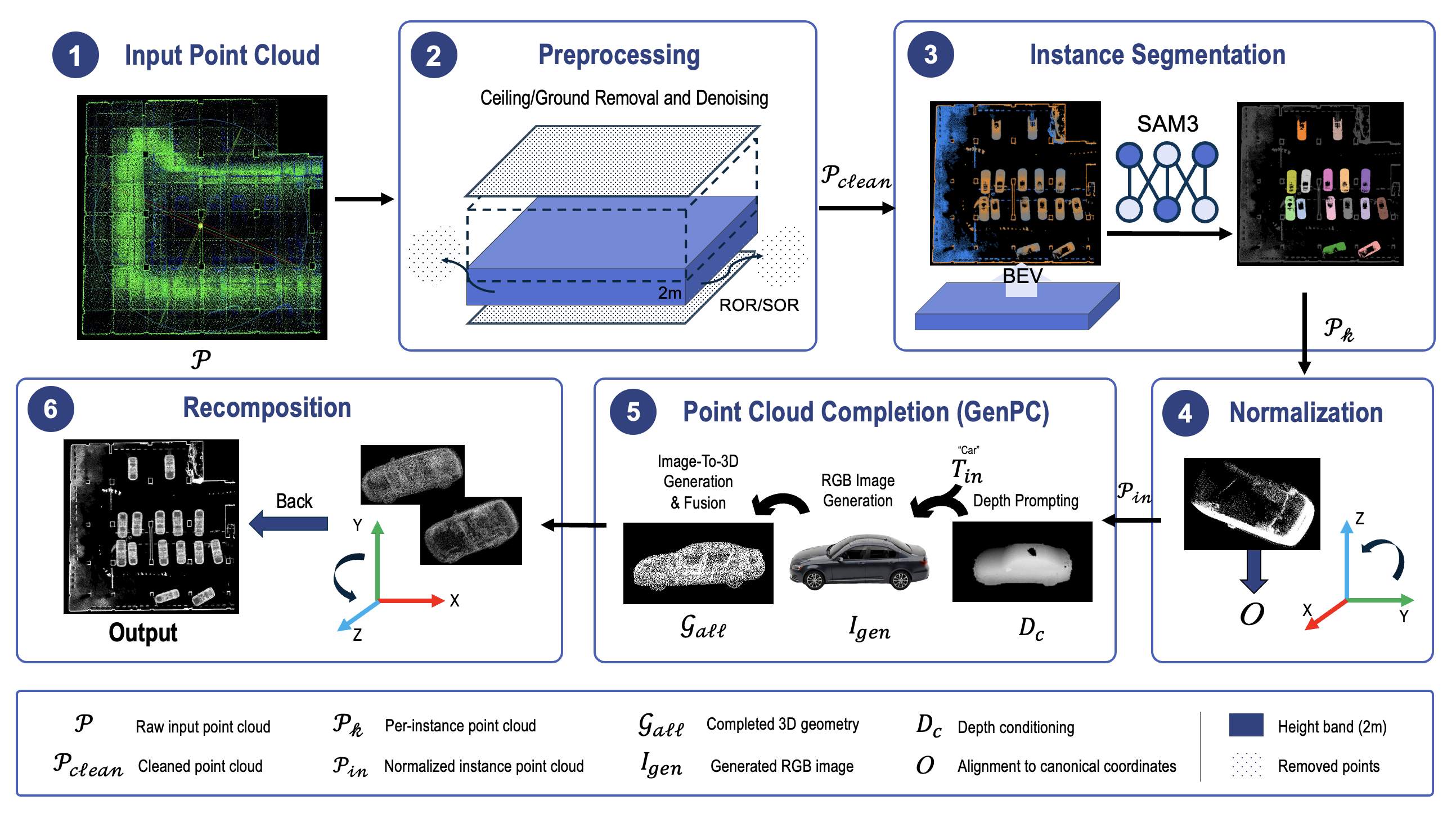}
    \caption{Overview of the proposed ZVeC pipeline. The input point cloud is first preprocessed by removing the ceiling, ground, and noise before being partitioned into individual vehicle instances using a Segment Anything Model-based instance segmentation module. Each normalized instance is then completed using our GenPC-based completion module, after which the reconstructed instances are transformed back and reinserted into the original scene to produce the final completed point cloud.}
    \label{fig: pipeline}
\end{figure}

Underground parking environments present a particularly challenging setting because severe occlusions and restricted viewpoints make complete supervision prohibitively expensive to obtain. Nevertheless, parked vehicles exhibit strong category-level geometric regularities despite substantial missing observations. This suggests that scene completion need not be formulated as a global reconstruction problem. Instead, decomposing the scene into semantic object instances transforms completion into independent geometric inference over a low-dimensional object shape space, reducing reconstruction ambiguity while enabling zero-shot completion through generalized geometric priors rather than scene-specific supervision.

Motivated by this observation, we propose \textbf{ZVeC}, a zero-shot, instance-driven framework for point cloud completion in real-world LiDAR scenes. Rather than learning a scene-specific reconstruction model, ZVeC completes each object independently using our proposed \textbf{GenPC} module, which recovers missing geometry from partial observations without additional training or fine-tuning. The framework first extracts vehicle instances using a Segment Anything Model-based segmentation module, applies zero-shot completion to each normalized instance, and finally recomposes the reconstructed objects into the original scene while preserving the global spatial layout.

This formulation offers three key advantages. First, instance-level decomposition reduces interference from neighboring objects and background clutter. Second, zero-shot completion improves generalization across diverse environments by avoiding scene-specific supervision. Third, the recovered instance representations naturally support downstream applications including digital twin construction, semantic mapping, and scene understanding.

To evaluate our approach, we construct a dense real-world LiDAR benchmark collected in underground parking environments. Although the scans contain high point densities, severe occlusions and limited viewpoints create substantial missing geometry, providing a realistic benchmark for zero-shot completion. Experimental results demonstrate consistent improvements over representative scene-level baselines in both quantitative metrics and visual quality, particularly in recovering fine-grained vehicle structures.

\noindent\textbf{Contributions.}
\begin{itemize}
    \item We propose \textbf{ZVeC}, a zero-shot framework that reformulates scene-level point cloud completion as compositional instance-level reconstruction, reducing reconstruction ambiguity in cluttered LiDAR scenes through object-wise completion with the proposed \textbf{GenPC} module.
    
    \item We introduce a real-world underground parking LiDAR benchmark for evaluating zero-shot point cloud completion under realistic occlusions and incomplete observations.\footnote{Our dataset will be released after double blind peer-review.}
    
    \item Extensive experiments demonstrate consistent improvements over representative baselines and robust reconstruction under extreme input sparsity, maintaining geometrically consistent completions using only 1\% of the original input points.
\end{itemize}


\section{Related Work}

\subsection{Instance-level Point Cloud Completion}

Instance-level point cloud completion aims to recover complete object geometry from sparse and partial observations~\cite{b9,b10}. Early learning-based methods, including PCN~\cite{PCN} and GRNet~\cite{GRNet}, established encoder-decoder architectures for directly reconstructing complete shapes from partial point clouds. Recent advances have shifted the field toward Transformer-based and generative models~\cite{b6,b9,b10,b11}.

Transformers improve completion by modeling long-range geometric dependencies through self-attention~\cite{b6}. PoinTr~\cite{b6} formulates completion as a set-to-set translation task by representing point clouds as geometry-aware point proxies within a Transformer encoder-decoder. AdaPoinTr~\cite{b10} extends this framework with adaptive query generation and multi-scale geometric feature extraction. To better recover local geometry, SnowflakeNet~\cite{b12} introduces a Skip-Transformer with progressive point splitting, while SeedFormer~\cite{b13} employs hierarchical ``Patch Seeds'' for coarse-to-fine shape generation.

Despite their strong benchmark performance, these methods rely primarily on geometric observations, resulting in ambiguous predictions under severe occlusions or out-of-distribution inputs. Recent work addresses this limitation by incorporating generative image priors. SDS-Complete~\cite{b11} performs test-time optimization using Score Distillation Sampling and a pretrained text-to-image diffusion model, enabling zero-shot completion without large-scale 3D supervision at significant computational cost. In contrast, GenPC~\cite{b14} leverages feed-forward image-to-3D generative priors with depth prompting and geometry-preserving fusion, enabling efficient zero-shot completion of real-world scans without iterative optimization while conditioning on depth observations from the input point cloud.

\subsection{3D Instance Segmentation}

Recent 3D indoor instance segmentation methods can be broadly categorized into native 3D approaches and 2D-to-3D lifting methods. Early native 3D methods, including GSPN~\cite{GSPN}, 3D-BoNet~\cite{BoNet}, and OccuSeg~\cite{OccuSeg}, introduced proposal-based and occupancy-aware paradigms for directly segmenting object instances from point clouds. More recent bottom-up approaches, such as PointGroup~\cite{b15}, HAIS~\cite{HAIS}, and SoftGroup~\cite{b16}, progressively improved instance grouping through point clustering and hierarchical aggregation, while Transformer-based methods such as Mask3D~\cite{Mask3D} directly predict instance masks using learnable object queries. Although highly effective, these native 3D methods remain limited by severe occlusions and the scarcity of densely annotated 3D datasets.

To reduce dependence on large-scale 3D annotations, 2D-to-3D lifting methods leverage mature 2D vision models~\cite{b17,b18}. These approaches first segment objects in multi-view RGB-D images before projecting the resulting masks into 3D using camera geometry~\cite{b19,b20}. PanopticFusion~\cite{b21} fuses 2D instance predictions into a volumetric TSDF representation, while 3D-SIS~\cite{b22} lifts high-resolution RGB features to improve 3D instance segmentation. A key challenge in 2D-to-3D lifting is maintaining multi-view consistency, as objects are observed from varying viewpoints, scales, and occlusion patterns. Recent methods~\cite{b18,b23} address this using spatial-temporal fusion, bipartite graph matching, or 3D consensus clustering to merge redundant predictions into coherent object instances. Following this paradigm, we perform instance segmentation in Bird's-Eye View (BEV) space and exploit the zero-shot capability of Segment Anything Model 3 (SAM3)~\cite{b24} to obtain accurate instance boundaries.

\section{Methodology}


The framework starts from a dense point cloud $\mathcal{P}$, where ceiling and ground removal with denoising are first applied (Figure~\ref{fig: pipeline}). The cleaned cloud is projected to a 2D bird's-eye view for zero-shot instance segmentation using SAM3~\cite{b24}. The resulting masks are back-projected to 3D to extract clusters $\mathcal{P}_k$, followed by geometric filtering to remove artifacts. After normalization, each partial vehicle cloud is processed by GenPC~\cite{b14}, which projects $\mathcal{P}_{in}$ to 2D and synthesizes an RGB image $I_{gen}$ through depth inpainting $D_c$ for initial 3D completion. The generated shape is aligned with the original scan and represented as 3D Gaussians $\mathcal{G}_{all}$, where missing regions are refined using Score Distillation Sampling while observed geometry remains fixed before completed instances are reinserted into the scene.

\subsection{Data Preprocessing}

To efficiently process large-scale parking lot point clouds, we introduce a multi-stage preprocessing pipeline that reduces computational cost. Exploiting the prior that the ceiling and ground form near-parallel horizontal planes, we perform $Z$-axis histogram slicing followed by RANSAC plane fitting on the highest and lowest slices to estimate the boundary planes. A gravity-axis height filter then removes ground points and points exceeding a vehicle height of $2\,\mathrm{m}$.

Residual noise is removed using a decoupled downsampling and mapping strategy. Radius Outlier Removal (ROR) and Statistical Outlier Removal (SOR) are applied only to a voxel-downsampled subset. Detected outliers are mapped back to the original dense cloud and removed using bulk radius queries with a $k$-d tree. The resulting point cloud $\mathcal{P}_{clean}$ is defined as
\begin{equation}
    \mathcal{P}_{clean} = \left\{ p \in \mathcal{P}_{raw} \mid \exists q \in \mathcal{D}_{\text{outlier}}, \| p - q \|_2 \le r \right\},
\end{equation}
where $\mathcal{P}_{raw}$ is the original point cloud, $\mathcal{D}_{\text{outlier}}$ is the detected outlier set, and $r$ is the expansion radius. This vectorized mapping avoids serial point-wise operations while maintaining identical filtering quality and accelerating denoising by up to two orders of magnitude.

\subsection{Instance Segmentation}

To perform 3D instance segmentation, we adapt SAM3~\cite{b24} to point clouds through a Bird's-Eye View (BEV) projection. Given the input point cloud $\mathcal{P}_{clean}=\{p_i=(x_i,y_i,z_i)\}_{i=1}^{N}$, we discretize the horizontal plane into a 2D grid with resolution $\Delta r$. To preserve height information after projection, the vertical coordinates are normalized ($\bar{z}\in[0,1]$) and encoded using a fixed pseudo-color mapping ($R=\bar{z}$, $G=0.5$, $B=1-\bar{z}$). During rasterization, we cache a sparse inverse mapping from each BEV pixel to its corresponding 3D point indices. SAM3 predicts a set of binary masks $\{M_k\}$, which are projected back into 3D through the cached mapping to produce segmented point cloud instances $\mathcal{P}_k \subset \mathcal{P}_{clean}$.

To suppress false positives introduced by the top-down projection, each lifted instance is filtered using simple geometric constraints. For an instance $\mathcal{P}_k$, we compute its bounding-box height $H_k$, maximum horizontal length $L_k$, vertical standard deviation $\sigma_z$, and the fraction of points above ground level $\rho_k$. An instance is retained only if

\begin{equation}
\begin{split}
\mathcal{F}(\mathcal{P}_k)=\mathbb{I}(&|\mathcal{P}_k|\ge n_{\min}\land h_{\min}\le H_k\le h_{\max}\\
&\land L_k\le l_{\max}\land \rho_k\ge\tau_{\mathrm{ratio}}\land \sigma_z\ge\tau_{\mathrm{std}}),
\end{split}
\label{eq:geo_filter}
\end{equation}
where $n_{\min}$ is the minimum number of points, $h_{\min}$ and $h_{\max}$ define the allowable vehicle height range, $l_{\max}$ is the maximum allowable vehicle length, $\tau_{\mathrm{ratio}}$ is the minimum fraction of elevated points, and $\tau_{\mathrm{std}}$ is the minimum vertical standard deviation required to reject nearly planar structures. Instances that fail these criteria are discarded as projection artifacts. Although similar filtering could be achieved by increasing the confidence thresholds of SAM3, doing so often suppresses heavily occluded vehicles with incomplete geometry.

\subsection{Canonical Instance Normalization}

\begin{table}[t]
\centering
\caption{Summary statistics of the completed point clouds over all 15 vehicle instances. Values are reported as mean $\pm$ standard deviation across instances. Lower is better for all metrics except voxel coverage.}
\label{tab:metrics}
\begin{tabular}{lccc}
\toprule
\textbf{Metric} & \textbf{Mean $\pm$ Std.} & \textbf{Min} & \textbf{Max} \\
\midrule
Surface Smoothness $\downarrow$ & $0.010 \pm 0.001$ & 0.009 & 0.012 \\
Density CV $\downarrow$         & $0.109 \pm 0.003$ & 0.105 & 0.115 \\
Voxel Coverage $\uparrow$       & $0.161 \pm 0.021$ & 0.123 & 0.188 \\
\midrule
KL Divergence $\downarrow$      & $2.121 \pm 0.359$ & 1.760 & 3.148 \\
EMD $\downarrow$                & $0.239 \pm 0.045$ & 0.159 & 0.329 \\
Sinkhorn Distance $\downarrow$  & $0.288 \pm 0.044$ & 0.211 & 0.375 \\
L1 Distance $\downarrow$        & $0.094 \pm 0.018$ & 0.058 & 0.120 \\
L2 Distance $\downarrow$        & $0.143 \pm 0.036$ & 0.076 & 0.198 \\
\bottomrule
\end{tabular}
\end{table}

Following instance segmentation, each object is transformed into a canonical coordinate frame before point cloud completion. This normalization standardizes the coordinate system, pose, and scale, ensuring consistent depth rendering for the completion model while recording the transformations required for later scene reconstruction.

Let the segmented point cloud be
\begin{equation}
\mathcal{P}=\{\mathbf{p}_i\}_{i=1}^{N},
\qquad
\mathbf{p}_i=[x_i,y_i,z_i]^\top.
\end{equation}

If the input is represented in a Z-up coordinate system, it is first converted to a standard Y-up right-handed frame,
\begin{equation}
\mathbf{p}_i^{(y)}
=
\mathbf{R}_{z\rightarrow y}\mathbf{p}_i^{(z)},
\qquad
\mathbf{R}_{z\rightarrow y}
=
\begin{bmatrix}
1 & 0 & 0\\
0 & 0 & 1\\
0 & -1 & 0
\end{bmatrix}.
\end{equation}

The point cloud is then centered at its centroid,
\begin{equation}
\tilde{\mathbf{p}}_i
=
\mathbf{p}_i^{(y)}
-
\mathbf{c},
\qquad
\mathbf{c}
=
\frac{1}{N}
\sum_{i=1}^{N}
\mathbf{p}_i^{(y)},
\end{equation}
which is more robust to irregular object geometry than axis-aligned bounding-box centering.

To estimate the object heading, the centered point cloud is projected onto the ground plane and Principal Component Analysis (PCA) is applied. Let
\begin{equation}
\mathbf{q}_i=[\tilde{x}_i,\tilde{z}_i]^\top,
\end{equation}
with covariance
\begin{equation}
\mathbf{\Sigma}
=
\frac{1}{N}
\sum_{i=1}^{N}
(\mathbf{q}_i-\bar{\mathbf{q}})
(\mathbf{q}_i-\bar{\mathbf{q}})^\top,
\qquad
\bar{\mathbf{q}}
=
\frac{1}{N}
\sum_{i=1}^{N}
\mathbf{q}_i.
\end{equation}
The dominant principal component defines the horizontal heading direction. For nearly symmetric or partially observed objects, PCA may produce a $90^\circ$ ambiguity. We resolve this by comparing the principal direction with the longer horizontal axis of the axis-aligned bounding box and rotating it by $90^\circ$ when inconsistent. The sign ambiguity is removed by enforcing a positive X direction.

The resulting heading angle $\theta$ defines the canonical rotation,
\begin{equation}
\theta
=
\operatorname{atan2}(v_z,v_x),
\qquad
\hat{\mathbf{p}}_i
=
\mathbf{R}_y(-\theta)\tilde{\mathbf{p}}_i,
\end{equation}
where
\begin{equation}
\mathbf{R}_y(\theta)
=
\begin{bmatrix}
\cos\theta & 0 & -\sin\theta\\
0 & 1 & 0\\
\sin\theta & 0 & \cos\theta
\end{bmatrix}.
\end{equation}

To obtain a unique canonical orientation, the aligned point cloud is required to satisfy
\begin{equation}
l_x\ge l_z.
\end{equation}
If this condition is violated, an additional rotation of $\pi/2$ about the Y-axis is applied. The rotated point cloud is then recentered using the axis-aligned bounding-box center,
\begin{equation}
\bar{\mathbf{p}}_i
=
\hat{\mathbf{p}}_i
-
\mathbf{o},
\qquad
\mathbf{o}
=
\frac{1}{2}
\begin{bmatrix}
\max \hat{x}+\min \hat{x}\\
\max \hat{y}+\min \hat{y}\\
\max \hat{z}+\min \hat{z}
\end{bmatrix}.
\end{equation}

Finally, the instance is normalized by its maximum bounding-box extent,
\begin{equation}
s
=
\max(d_x,d_y,d_z),
\qquad
\mathbf{p}_i^{\mathrm{norm}}
=
\frac{\bar{\mathbf{p}}_i}{s}.
\end{equation}

The resulting canonical representation is Y-up, centered, consistently oriented, and normalized to unit scale. The corresponding translation, rotation, coordinate transformation, and scale parameters are retained to restore the completed instances to the original scene coordinate frame during recomposition.

\subsection{Point Cloud Completion}

\begin{figure*}[t]
    \centering

    \begin{subfigure}{0.264\textwidth}
        \centering
        \includegraphics[width=\linewidth]{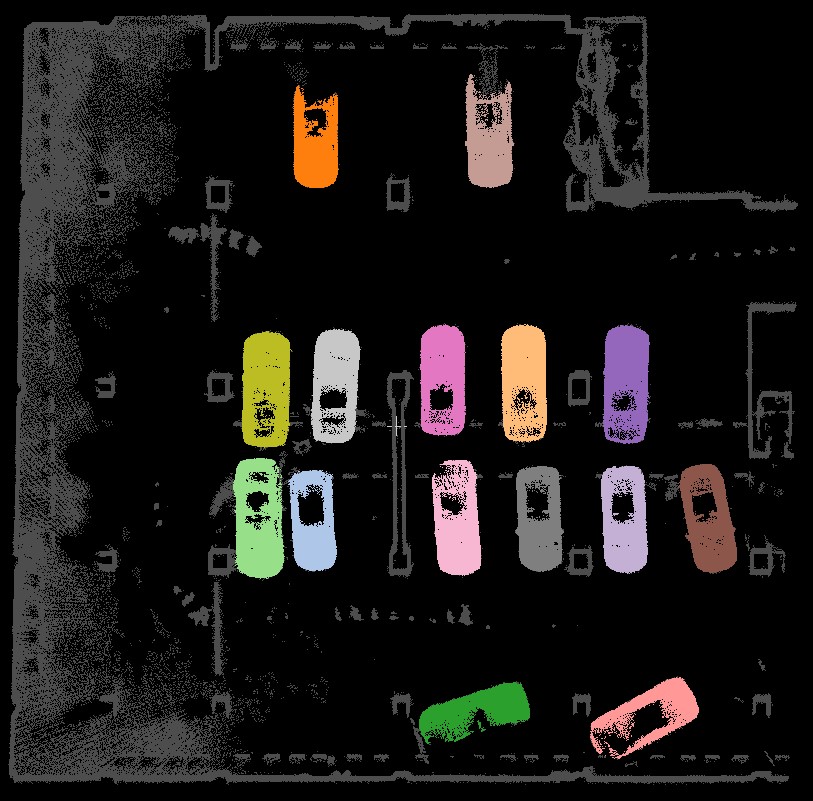}
        \caption{}
        \label{fig: VOIS}
    \end{subfigure}
    \hfill
    \begin{subfigure}{0.32\textwidth}
        \centering
        \includegraphics[width=\linewidth]{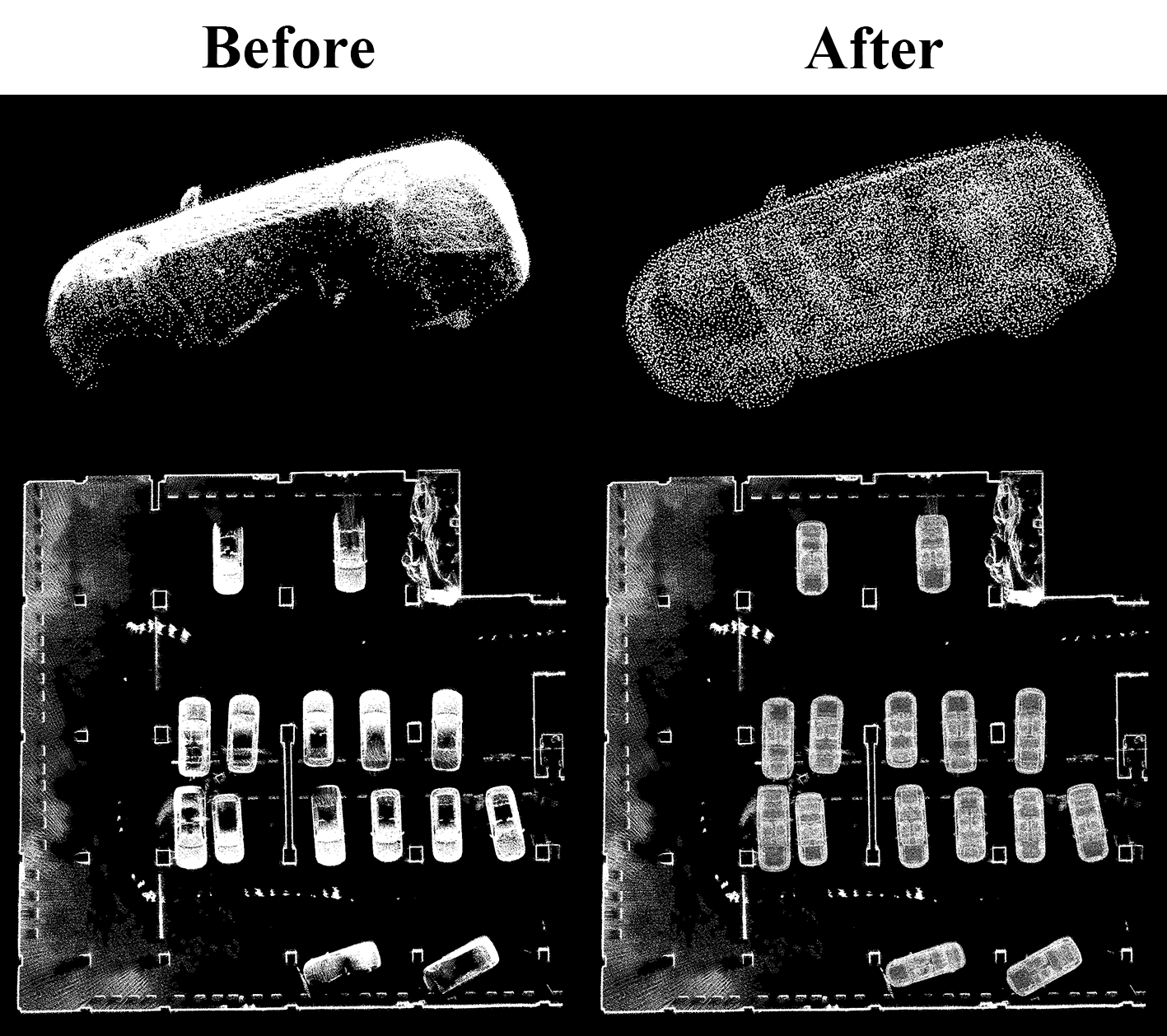}
        \caption{}
        \label{fig: VOC}
    \end{subfigure}
    \hfill
    \begin{subfigure}{0.33\textwidth}
        \centering
        \includegraphics[width=\linewidth]{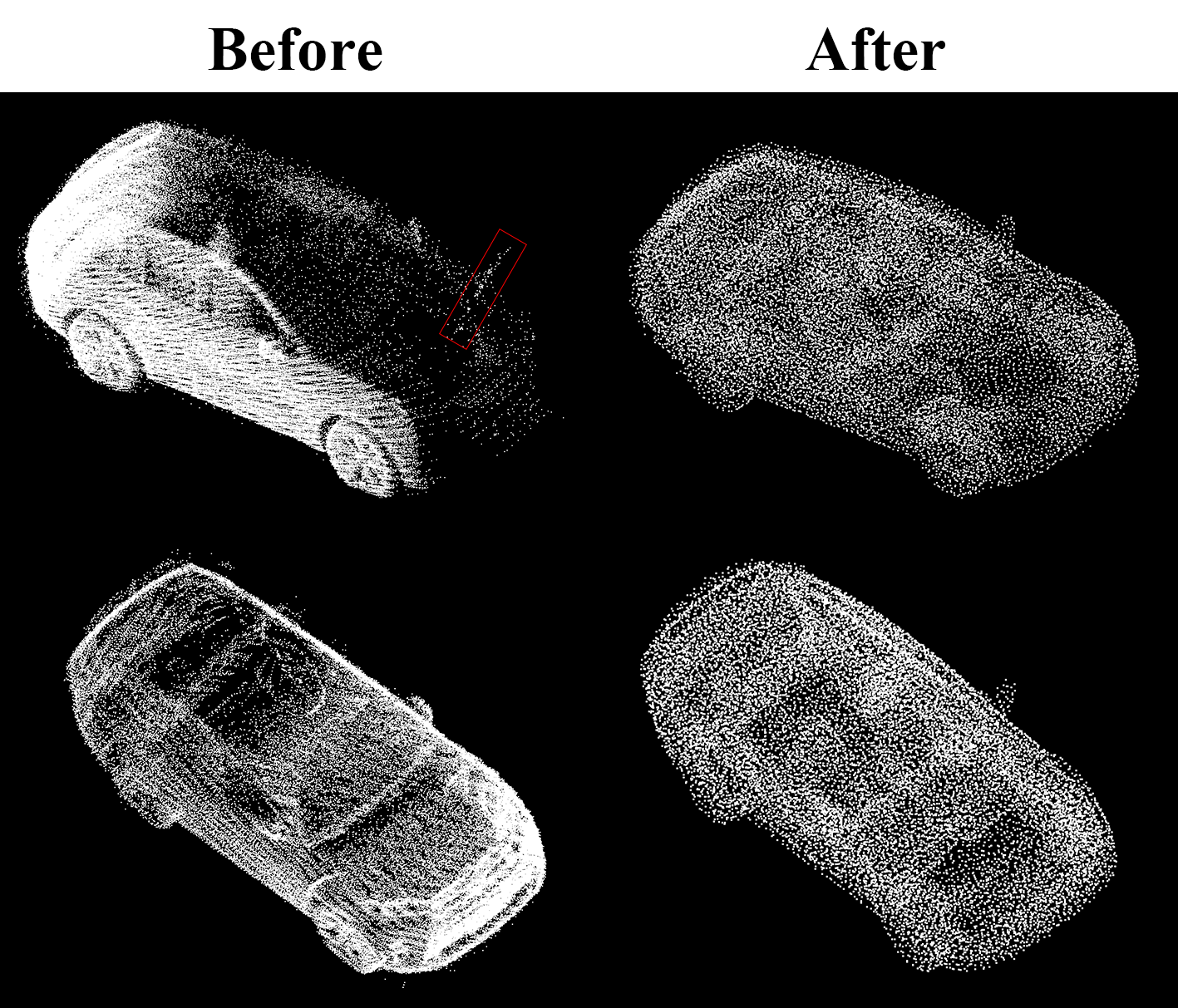}
        \caption{}
        \label{fig: ex}
    \end{subfigure}

    \caption{(\textbf{Left}) Scene decomposition via instance segmentation. Our pipeline first segments target objects into instances, then reconstructs each object's point cloud independently. (\textbf{Center}) Qualitative results of point cloud completion. Visualization of the initial and reconstructed point cloud datasets. A per-instance result is shown in the top row. (\textbf{Right}) Visual comparison of least and most changed instances by KL divergence. Top: most changed. Bottom: least changed. Large changes after completion reflect differences in the quality of the initial point cloud rather than the final reconstruction quality.}
\end{figure*}

Our GenPC-based completion module reconstructs complete vehicle geometries from sparse partial scans using the zero-shot framework of GenPC~\cite{b14}. Given a normalized point cloud $\mathcal{P}_{in}\subseteq\mathbb{R}^{N\times3}$ and semantic prompt $T_{in}$ (e.g., ``car''), the input is projected to a coarse depth map $D_{raw}$. A diffusion-based depth inpainting model produces a completed depth map $D_c$, which conditions a ControlNet to synthesize an RGB image $I_{gen}$. The generated image is then converted into an initial complete point cloud $\mathcal{P}_{gen}$ using a pre-trained image-to-3D model.

To preserve the observed geometry, the input point cloud is colorized to form $\mathcal{P}_{partial}$ and aligned with $\mathcal{P}_{gen}$ by optimizing a scale factor $s$ that minimizes geometric and color-based Chamfer Distances:
\begin{equation}
\begin{split}
s^*=\arg\min_s\Big(
&\alpha\,CD_{XYZ}(\mathcal{P}_{partial},s\mathcal{P}_{gen})\\
&+\beta\,CD_{RGB}(\mathcal{P}_{partial},s\mathcal{P}_{gen})
\Big).
\end{split}
\label{eq:optimization}
\end{equation}

The aligned point cloud is represented as 3D Gaussians $\mathcal{G}_{all}$, where the observed geometry $\mathcal{G}_{partial}$ remains fixed while the missing regions $\mathcal{G}_{miss}$ are refined using Score Distillation Sampling (SDS). This optimization preserves the measured vehicle geometry while improving the fidelity of the synthesized regions.


\subsection{Recomposition}

After completion, each object resides in a normalized Y-up coordinate system. To recover the original scene coordinates, the inverse preprocessing transformations are applied to the completed point cloud
\begin{equation}
\mathcal{P}_{\mathrm{comp}}
=
\left\{
\mathbf{p}_i^{\mathrm{comp}}
\right\}_{i=1}^{M}.
\end{equation}

If heading normalization was applied during preprocessing, each point is restored by
\begin{equation}
\mathbf{p}_i^{\mathrm{scene}}
=
\mathbf{R}_{z\rightarrow y}^{\top}
\left(
\mathbf{R}_y(\theta)^{\top}
\left(
s\mathbf{p}_i^{\mathrm{comp}}
+
\mathbf{o}
\right)
+
\mathbf{b}
\right),
\end{equation}
where $s$, $\mathbf{o}$, $\theta$, and $\mathbf{b}$ denote the stored scale factor, post-rotation offset, heading angle, and bounding-box center, respectively. When heading normalization or axis conversion is not applied, the corresponding transformations reduce to the identity.

For $K$ completed instances, the restored object set is
\begin{equation}
\mathcal{P}_{\mathrm{obj}}
=
\bigcup_{k=1}^{K}
\mathcal{P}_k,
\end{equation}
and the reconstructed scene is obtained by merging the restored objects with the background point cloud,
\begin{equation}
\mathcal{P}_{\mathrm{scene}}
=
\mathcal{P}_{\mathrm{bg}}
\cup
\bigcup_{k=1}^{K}
\mathcal{P}_k.
\end{equation}

\section{Experiments and Analysis}

\begin{figure*}[t]
    \centering
    \begin{subfigure}{0.49\linewidth}
        \centering
        \includegraphics[width=\linewidth]{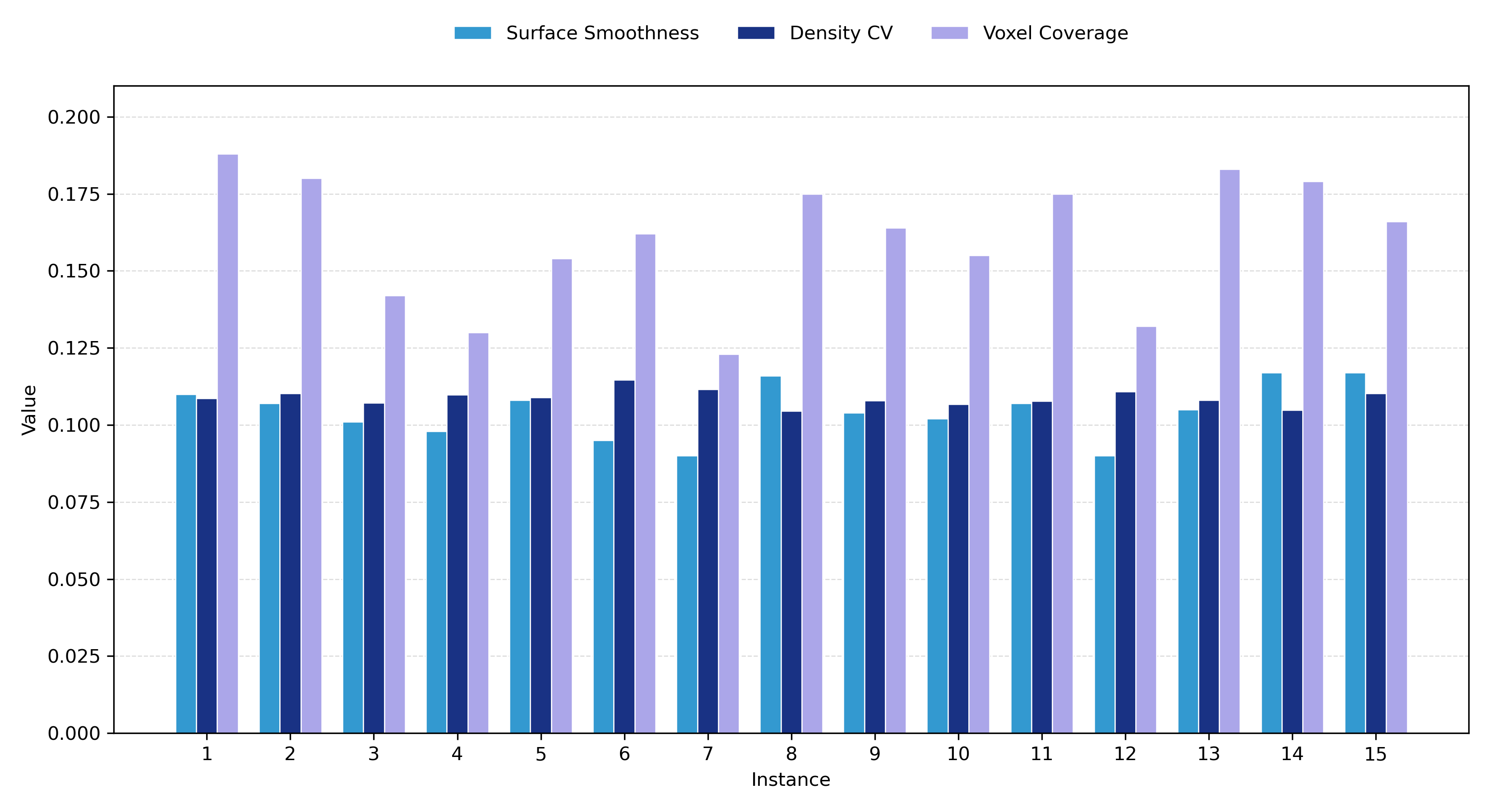}

    \end{subfigure}
    \hfill
    \begin{subfigure}{0.49\linewidth}
        \centering
        \includegraphics[width=\linewidth]{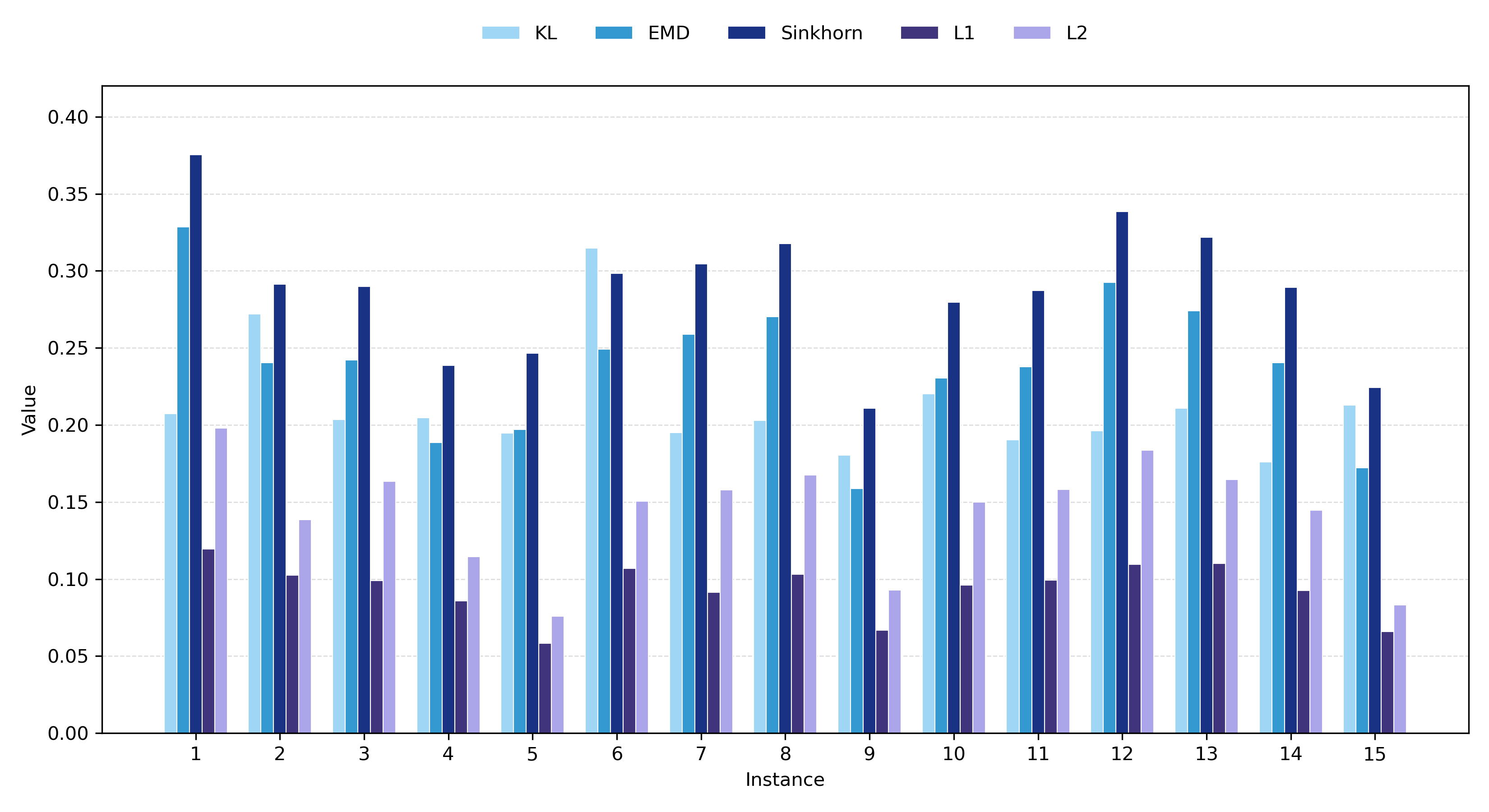}

    \end{subfigure}

    \caption{Histograms of the evaluation metrics. (Left) No-reference geometry-based metrics show consistent geometric quality across the final per-instance point clouds. (Right) Full-reference metrics exhibit substantial variation in the pre- and post-completion differences across instances, primarily due to differences in the quality and completeness of the initial point clouds.}\label{fig:fr} \label{fig:nr}
    \label{fig:histograms}
\end{figure*}
All experiments were conducted using two computational environments. The first two stages were performed on a local computer equipped with an Intel Core i7-9700K CPU running at 3.60 GHz, 64 GB of RAM, and an NVIDIA GeForce RTX 2080 Ti GPU with 11 GB of memory, executed in Docker Desktop using the pytorch/pytorch:2.7.0-cuda12.6-cudnn9-devel container image, with Ubuntu 22.04 as the runtime environment.

The rest of the stages were executed on a remote GPU server. The server was configured with an NVIDIA Triton Server 24.12 image, Python 3.12, Ubuntu 24.04, and CUDA 12.6. Its hardware configuration consisted of 16 vCPUs based on an Intel Xeon Gold 6430 processor, 120 GB of RAM, and a single NVIDIA GeForce RTX 4090 GPU with 24 GB of memory.

\subsection{Dataset}

We validated our pipeline on our newly collected real-world dense LiDAR point cloud dataset of an underground parking lot using an Orbis mobile scanning system. The acquisition was conducted over an approximately 8-minute scanning sequence, capturing dense LiDAR observations of parked vehicles, structural elements, and surrounding indoor geometry. Based on a robust estimate of the horizontal spatial extent of the reconstructed point cloud, the scanned area covers approximately \(36\mathrm{m} \times 40\mathrm{m}\). The per-point attributes in our dataset comprise spatial coordinates (x, y, z), intensity, and GPS time. Despite the relatively high density of the scan, localized data gaps and geometric incompleteness remain due to occlusions and limited viewpoints in a realistic single-pass drive through.

\subsection{Evaluation Metrics}

Since complete ground-truth geometry is unavailable for real-world LiDAR scans, conventional full-reference evaluation cannot be directly applied. We therefore assess reconstruction quality using (i) no-reference geometric metrics that quantify the quality of the completed point cloud and (ii) distribution- and distance-based metrics that measure consistency between the completed point cloud $\mathcal{P}_{g}$ and the observed input $\mathcal{P}_{r}$.

\subsubsection{No-Reference Metrics}

The following metrics evaluate the geometric quality of the completed point cloud $\mathcal{P}_{g}$ without requiring ground truth.

\begin{itemize}

\item \textbf{Surface Smoothness ($S_{sm}$):}
Measures local planarity by averaging the orthogonal distance from each point to its local tangent plane:
\begin{equation}
S_{sm}=\frac{1}{|\mathcal{P}_{g}|}\sum_{p_i\in\mathcal{P}_{g}}
\left|(p_i-\mu_i)\cdot n_i\right|,
\end{equation}
where $n_i$ and $\mu_i$ denote the local surface normal and neighborhood centroid, respectively.

\item \textbf{Density Coefficient of Variation ($D_{cv}$):}
Measures spatial density uniformity:
\begin{equation}
D_{cv}=\frac{\sigma_\rho}{\mu_\rho+\epsilon},
\end{equation}
where $\mu_\rho$ and $\sigma_\rho$ are the mean and standard deviation of the local point density.

\item \textbf{Voxel Coverage ($V_{cov}$):}
Measures spatial occupancy by computing the fraction of occupied voxels within the discretized 3D bounding volume.

\end{itemize}

\subsubsection{Distribution-Based Metrics}

The following metrics quantify the similarity between the completed point cloud $\mathcal{P}_{g}$ and the observed point cloud $\mathcal{P}_{r}$.

\begin{itemize}

\item \textbf{KL Divergence ($D_{KL}$):}
Measures the divergence between 3D occupancy histograms:
\begin{equation}
D_{KL}(\mathcal{P}_{r}\|\mathcal{P}_{g})
=
\sum_x
H_r(x)
\log
\frac{H_r(x)}
{H_g(x)+\epsilon},
\end{equation}
where $H_r$ and $H_g$ denote 3D histograms computed over a common bounding box.

\item \textbf{Earth Mover's Distance (EMD):}
Measures the optimal transport cost between equally sampled point sets:
\begin{equation}
\mathrm{EMD}(\mathcal{P}_{g},\mathcal{P}_{r})
=
\min_{\phi}
\frac{1}{N}
\sum_{x\in\mathcal{P}_{g}}
\|x-\phi(x)\|_2,
\end{equation}
where $\phi$ is the optimal bijection.

\item \textbf{Sinkhorn Distance ($d_{\mathcal S}$):}
Approximates EMD using entropy-regularized optimal transport:
\begin{equation}
d_{\mathcal S}
=
\sum_{i,j}
P_{ij}
\|x_i-y_j\|_2.
\end{equation}

\item \textbf{Chamfer L1 ($d_{L1}$) and L2 ($d_{L2}$):}
Measure symmetric nearest-neighbor distances between $\mathcal{P}_{g}$ and $\mathcal{P}_{r}$. Let
\begin{equation}
\Lambda=
\left\{
\min_{y\in\mathcal P_r}\|x-y\|_2
\right\}_{x\in\mathcal P_g}
\cup
\left\{
\min_{x\in\mathcal P_g}\|y-x\|_2
\right\}_{y\in\mathcal P_r},
\end{equation}
then
\[
d_{L1}=\mathbb E_{d\in\Lambda}[|d|],
\qquad
d_{L2}=
\sqrt{\mathbb E_{d\in\Lambda}[d^2]}.
\]

\end{itemize}

\subsection{Point Cloud Completion Results}

\begin{table}[t]
\centering
\caption{Incremental ablation of the proposed normalization pipeline. Lower is better.}
\label{tab: ab}
\begin{tabular}{lcccc}
\toprule
\textbf{Configuration} &
\rotatebox{90}{\textbf{CD-L1}} &
\rotatebox{90}{\textbf{CD-L2}} &
\rotatebox{90}{\textbf{EMD}} &
\rotatebox{90}{\textbf{Sinkhorn}} \\
\midrule
No normalization     & 9.28 & 450.74 & 18.22 & 67.68 \\
+ Centroid centering & 0.418 & 0.991 & 0.741 & 0.855 \\
+ Orientation adjustment    & 0.127 & 0.051 & 0.148 & 0.094 \\
Ours  & \textbf{0.020} & \textbf{0.002} & \textbf{0.044} & \textbf{0.030} \\
\bottomrule
\end{tabular}
\end{table}

\begin{figure*}[t]
    \centering
    \includegraphics[width=1\linewidth]{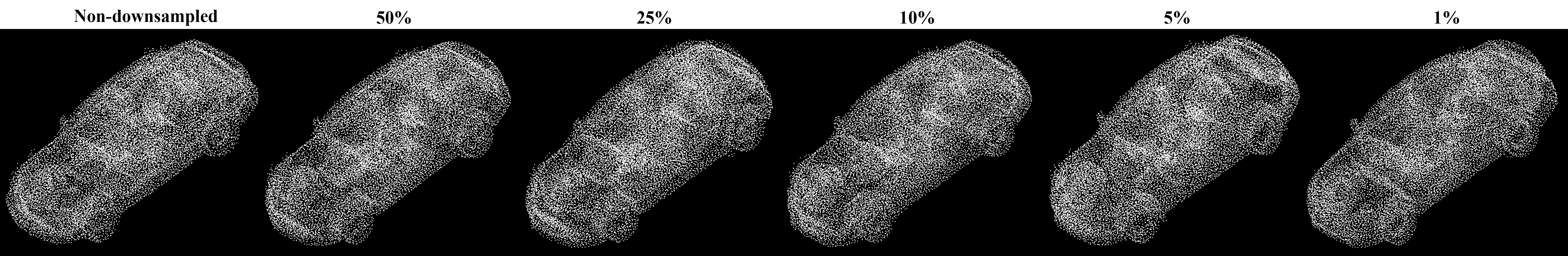}
\caption{Reconstruction results for a representative vehicle instance under varying levels of input sparsity. Despite significant differences in the number of input points, the completed geometry remains largely unchanged, demonstrating the robustness of the method to input sparsity.}
    \label{fig: VOS}
\end{figure*}

The quantitative results demonstrate consistent reconstruction quality across all 15 vehicle instances. For the no-reference metrics in Table~\ref{tab:metrics} and Figure~\ref{fig:nr} (surface smoothness scaled by $\times10$ for visualization), surface smoothness ($0.010 \pm 0.001$) and density CV ($0.109 \pm 0.003$) exhibit very small variance, indicating smooth surfaces and uniform point sampling across the completed vehicles. Voxel coverage averages $0.161 \pm 0.021$, reflecting expected variation in the amount of recovered geometry due to differing occlusion levels.

For the full-reference metrics in Table~\ref{tab:metrics} and Figure~\ref{fig:fr} (KL divergence scaled by $\frac{1}{10}$), KL divergence averages $2.121 \pm 0.359$, indicating varying degrees of point density redistribution during completion. In contrast, EMD ($0.239 \pm 0.045$), Sinkhorn distance ($0.288 \pm 0.044$), and the Chamfer L1 ($0.094 \pm 0.018$) and L2 ($0.143 \pm 0.036$) distances remain consistently low with relatively small variance, demonstrating stable geometric alignment between the completed and observed point clouds.

Unlike EMD, Sinkhorn, and Chamfer distances, which measure geometric correspondence, KL divergence quantifies changes in the underlying point density distribution and is therefore more sensitive to the redistribution of probability mass introduced during completion. We consequently use KL divergence to guide the qualitative analysis while reporting all metrics.

This behavior is illustrated by Instances~6 and~14 in Figure~\ref{fig: ex}. Instance~6 exhibits the highest KL divergence due to severe occlusions, density imbalance, and segmentation artifacts (red box), requiring substantial redistribution of point density during completion. Nevertheless, the reconstructed geometry remains visually coherent, consistent with the stable no-reference and geometric correspondence metrics. In contrast, Instance~14 exhibits the lowest KL divergence because the input already closely approximates the underlying vehicle geometry, requiring only minor refinement during completion.

\begin{table*}[t]
\centering
\caption{Average KL divergence values under different downsampling ratios. Lower values indicate greater similarity.} \label{tab: kl} 
\begin{tabular}{lccccc}
\toprule
\textbf{Comparison} &
\textbf{50 \%} &
\textbf{25 \%} &
\textbf{10 \%} &
\textbf{5 \%} &
\textbf{1 \%} \\
\midrule
Completed (downsampled) vs. Raw (downsampled)
& 1.9717 & 2.3142 & 1.6554 & 1.4924 & 1.5233 \\

Completed (downsampled) vs. Raw (full)
& 2.2159 & 2.8865 & 2.3241 & 2.0662 & 2.3425 \\

Completed (downsampled) vs. Completed (full)
& 0.2904 & 0.3153 & 0.3423 & 0.3082 & 0.4955 \\
\bottomrule
\end{tabular}
\end{table*}

\subsection{Ablation Study}

To validate instance-to-scene grounding, we perform an ablation with four variants that progressively incorporate normalization and grounding. A near-complete instance is selected from the raw point cloud as the reference, and a partial input is generated by removing surface regions before completion. Performance is evaluated against the reference using symmetric Chamfer Distance (L1/L2), EMD, and Sinkhorn Distance.

As shown in Table~\ref{tab: ab}, removing normalization and grounding substantially degrades performance, confirming their importance. Centering alone provides only marginal improvement, while omitting the coordinate transformation further reduces accuracy, demonstrating that Z-up to Y-up alignment is essential. The full normalization pipeline consistently achieves the best performance.

\subsection{Robustness Analysis}

To evaluate robustness under varying observation sparsity, we randomly downsample the input point clouds to retention ratios of 50\%, 25\%, 10\%, 5\%, and 1\%. KL divergence measures the difference between completions generated from downsampled and full-resolution inputs, as well as between the completed and raw point clouds.

As shown in Table~\ref{tab: kl}, the completed point clouds remain highly consistent with the full-resolution completions. The KL divergence increases only from 0.2904 at 50\% retention to 0.4955 at 1\% retention, while Figure~\ref{fig: VOS} shows strong structural consistency across all sparsity levels. In contrast, the divergence between the completed and raw point clouds is substantially larger, reflecting the incompleteness of the observations. These results demonstrate that the proposed completion framework is robust to extreme input sparsity, producing consistent geometries even when only 1\% of the original points are retained.

\subsection{Discussion}
Our newly collected dataset lacks complete ground truth, so evaluation reflects a realistic deployment setting rather than exact reconstruction accuracy. We therefore assess per-instance geometric quality using no-reference metrics, quantify geometric and distributional changes before and after completion, and evaluate robustness under varying input sparsity. Manual inspection further confirms the plausibility and structural consistency of all completed instances. Future work will investigate automated point cloud quality assessment metrics for scalable evaluation.

Although this study focuses on vehicle reconstruction in indoor parking environments, the framework is largely environment- and class-agnostic. Future work will evaluate generalization to outdoor scenes and additional object categories, including pedestrians, traffic signs, and autonomous driving datasets, while introducing task-specific adaptations where necessary.

This work also advances object-centric digital twin representations. Future research will investigate object-level meshing and scene graph-based modeling with per-object pose, semantic attributes, and geometric descriptors for digital twin creations.

\section{Conclusions}

In this paper, we propose ZVeC, a zero-shot, instance-driven framework for point cloud completion in underground parking environments, following a compositional scene completion pipeline. By transforming a complex scene-level problem into instance-level reconstruction tasks, our method effectively overcomes the geometric incompleteness in point clouds caused by occlusions and clutter. Experiments on a real-world dense LiDAR dataset demonstrate that our framework achieves high-fidelity reconstructions both qualitatively and quantitatively. 

Beyond this specific scenario, ZVeC highlights the effectiveness of a zero-shot, instance-driven formulation for handling incomplete observations in structured environments. This paradigm provides a scalable and generalizable direction for robust 3D scene reconstruction and can benefit a range of downstream perception tasks, such as intelligent object-based scene representation for autonomous driving, embodied systems, and digital twin creation.

\bibliography{aaai2027}

@article{b1,
author  = "Guo, Yulan and Wang, Hanyun and Hu, Qingyong and Liu, Hao and Liu, Li and Bennamoun, Mohammed",
year    = 2021,
title   = "{Deep Learning for 3D Point Clouds: A Survey}",
journal = "IEEE Transactions on Pattern Analysis and Machine Intelligence",
volume  = 43,
number  = 12,
pages   = "4338--4364",
doi     = "10.1109/TPAMI.2020.3005434",
}

@article{b2,
author  = "Cui, Yaodong and Chen, Ren and Chu, Wenbo and Chen, Long and Tian, Daxin and Li, Ying and Cao, Dongpu",
year    = 2022,
title   = "{Deep Learning for Image and Point Cloud Fusion in Autonomous Driving: A Review}",
journal = "IEEE Transactions on Intelligent Transportation Systems",
volume  = 23,
number  = 2,
pages   = "722--739",
doi     = "10.1109/TITS.2020.3023541",
}

@inproceedings{b3,
author    = "Qin, Tong and Chen, Tongqing and Chen, Yilun and Shen, Shaojie",
year      = 2020,
title     = "{AVP-SLAM: Semantic Visual Mapping and Localization for Autonomous Vehicles in the Parking Lot}",
booktitle = "Proceedings of the IEEE/RSJ International Conference on Intelligent Robots and Systems {(IROS)}",
pages     = "5939--5945",
}

@article{b4,
author  = "Bresson, Guillaume and Alsayed, Zayed and Yu, Li and Glaser, S{'e}bastien",
year    = 2017,
title   = "{Simultaneous Localization and Mapping: A Survey of Current Trends in Autonomous Driving}",
journal = "IEEE Transactions on Intelligent Vehicles",
volume  = 2,
number  = 3,
pages   = "194--220",
doi     = "10.1109/TIV.2017.2749181",
}

@inproceedings{b5,
author    = "Yuan, Wentao and Khot, Tejas and Held, David and Mertz, Christoph and Hebert, Martial",
year      = 2018,
title     = "{PCN: Point Completion Network}",
booktitle = "Proceedings of the International Conference on 3D Vision {(3DV)}",
pages     = "728--737",
doi       = "10.1109/3DV.2018.00088",
}

@inproceedings{b6,
author    = "Yu, Xumin and Rao, Yongming and Wang, Ziyi and Liu, Zuyan and Lu, Jiwen and Zhou, Jie",
year      = 2021,
title     = "{PoinTr: Diverse Point Cloud Completion with Geometry-Aware Transformers}",
booktitle = "Proceedings of the IEEE/CVF International Conference on Computer Vision {(ICCV)}",
pages     = "12498--12507",
}

@inproceedings{b7,
author    = "Xie, Haozhe and Yao, Hongxun and Zhou, Shangchen and Mao, Jiageng and Zhang, Shengping and Sun, Wenxiu",
year      = 2020,
title     = "{GRNet: Gridding Residual Network for Dense Point Cloud Completion}",
booktitle = "Computer Vision -- ECCV 2020",
pages     = "365--381",
doi       = "10.1007/978-3-030-58545-7_21",
}

@article{b8,
author  = "Zhang, Kun and Zhang, Ao and Wang, Xiaohong and Li, Weisong",
year    = 2024,
title   = "{Deep-Learning-Based Point Cloud Completion Methods: A Review}",
journal = "Graphical Models",
volume  = 136,
pages   = "101233",
doi     = "10.1016/j.gmod.2024.101233",
}

@article{b9,
author  = "Chen, Zhikai and Long, Fuchen and Qiu, Zhaofan and Yao, Ting and Zhou, Wengang and Luo, Jiebo and Mei, Tao",
year    = 2024,
title   = "{Learning 3D Shape Latent for Point Cloud Completion}",
journal = "IEEE Transactions on Multimedia",
volume  = 26,
pages   = "8717--8729",
doi     = "10.1109/TMM.2024.3381814",
}

@article{b10,
author  = "Yu, Xumin and Rao, Yongming and Wang, Ziyi and Lu, Jiwen and Zhou, Jie",
year    = 2023,
title   = "{AdaPoinTr: Diverse Point Cloud Completion with Adaptive Geometry-Aware Transformers}",
journal = "IEEE Transactions on Pattern Analysis and Machine Intelligence",
volume  = 45,
number  = 12,
pages   = "14114--14130",
doi     = "10.1109/TPAMI.2023.3309253",
}

@inproceedings{b11,
author    = "Kasten, Yoni and Rahamim, Ohad and Chechik, Gal",
year      = 2023,
title     = "{Point Cloud Completion with Pretrained Text-to-Image Diffusion Models}",
booktitle = "Advances in Neural Information Processing Systems 36 {(NeurIPS)}",
pages     = "12171--12191",
url       = "https://proceedings.neurips.cc/paper_files/paper/2023/hash/284afdc2309f9667d2d4fb9290235b0c-Abstract-Conference.html",
}

@inproceedings{b12,
author    = "Xiang, Peng and Wen, Xin and Liu, Yu-Shen and Cao, Yan-Pei and Wan, Pengfei and Zheng, Wen and Han, Zhizhong",
year      = 2021,
title     = "{SnowflakeNet: Point Cloud Completion by Snowflake Point Deconvolution with Skip-Transformer}",
booktitle = "Proceedings of the IEEE/CVF International Conference on Computer Vision {(ICCV)}",
pages     = "5499--5509",
}

@inproceedings{b13,
author    = "Zhou, Haoran and Cao, Yun and Chu, Wenqing and Zhu, Junwei and Lu, Tong and Tai, Ying and Wang, Chengjie",
year      = 2022,
title     = "{SeedFormer: Patch Seeds Based Point Cloud Completion with Upsample Transformer}",
booktitle = "Computer Vision -- ECCV 2022",
pages     = "416--432",
doi       = "10.1007/978-3-031-20062-5_24",
}

@inproceedings{b14,
author    = "Li, An and Zhu, Zhe and Wei, Mingqiang",
year      = 2025,
title     = "{GenPC: Zero-Shot Point Cloud Completion via 3D Generative Priors}",
booktitle = "Proceedings of the IEEE/CVF Conference on Computer Vision and Pattern Recognition {(CVPR)}",
pages     = "1308--1318",
}

@inproceedings{b15,
author    = "Jiang, Li and Zhao, Hengshuang and Shi, Shaoshuai and Liu, Shu and Fu, Chi-Wing and Jia, Jiaya",
year      = 2020,
title     = "{PointGroup: Dual-Set Point Grouping for 3D Instance Segmentation}",
booktitle = "Proceedings of the IEEE/CVF Conference on Computer Vision and Pattern Recognition {(CVPR)}",
pages     = "4867--4876",
}

@inproceedings{b16,
author    = "Vu, Thang and Kim, Kookhoi and Luu, Tung M. and Nguyen, Thanh and Yoo, Chang D.",
year      = 2022,
title     = "{SoftGroup for 3D Instance Segmentation on Point Clouds}",
booktitle = "Proceedings of the IEEE/CVF Conference on Computer Vision and Pattern Recognition {(CVPR)}",
pages     = "2708--2717",
}

@inproceedings{b17,
author    = "Jatavallabhula, {Krishna Murthy} and Kuwajerwala, Alihusein and Gu, Qiao and Omama, Mohd and Chen, Tao and Li, Shuang and Iyer, Ganesh and Saryazdi, Soroush and Keetha, Nikhil and Tewari, Ayush and Tenenbaum, {Joshua B.} and {de Melo}, {Celso Miguel} and Krishna, Madhava and Paull, Liam and Shkurti, Florian and Torralba, Antonio",
year      = 2023,
title     = "{ConceptFusion: Open-Set Multimodal 3D Mapping}",
booktitle = "Robotics: Science and Systems {(RSS)}",
doi       = "10.15607/RSS.2023.XIX.066",
}

@inproceedings{b18,
author    = "Peng, Songyou and Genova, Kyle and Jiang, Chiyu {Max} and Tagliasacchi, Andrea and Pollefeys, Marc and Funkhouser, Thomas",
year      = 2023,
title     = "{OpenScene: 3D Scene Understanding with Open Vocabularies}",
booktitle = "Proceedings of the IEEE/CVF Conference on Computer Vision and Pattern Recognition {(CVPR)}",
pages     = "815--824",
}

@inproceedings{b19,
author    = "Miao, Xingyu and Duan, Haoran and Qian, Quanhao and Wang, Jiuniu and Long, Yang and Shao, Ling and Zhao, Deli and Xu, Ran and Zhang, Gongjie",
year      = 2025,
title     = "{Towards Scalable Spatial Intelligence via 2D-to-3D Data Lifting}",
booktitle = "Proceedings of the IEEE/CVF International Conference on Computer Vision {(ICCV)}",
pages     = "945--959",
}

@inproceedings{b20,
author    = "Yin, Yingda and Liu, Yuzheng and Xiao, Yang and Cohen-Or, Daniel and Huang, Jingwei and Chen, Baoquan",
year      = 2024,
title     = "{SAI3D: Segment Any Instance in 3D Scenes}",
booktitle = "Proceedings of the IEEE/CVF Conference on Computer Vision and Pattern Recognition {(CVPR)}",
pages     = "3292--3302",
}

@inproceedings{b21,
author    = "Narita, Gaku and Seno, Takashi and Ishikawa, Tomoya and Kaji, Yohsuke",
year      = 2019,
title     = "{PanopticFusion: Online Volumetric Semantic Mapping at the Level of Stuff and Things}",
booktitle = "Proceedings of the IEEE/RSJ International Conference on Intelligent Robots and Systems {(IROS)}",
pages     = "4205--4212",
doi       = "10.1109/IROS40897.2019.8967890",
}

@inproceedings{b22,
author    = "Hou, Ji and Dai, Angela and Nie{\ss}ner, Matthias",
year      = 2019,
title     = "{3D-SIS: 3D Semantic Instance Segmentation of RGB-D Scans}",
booktitle = "Proceedings of the IEEE/CVF Conference on Computer Vision and Pattern Recognition {(CVPR)}",
pages     = "4421--4430",
doi       = "10.1109/CVPR.2019.00455",
}

@inproceedings{b23,
  title={{Sam 3D: 3dfy anything in images}},
  author={Chen, Xingyu and Chu, Fu-Jen and Gleize, Pierre and Liang, Kevin J and Sax, Alexander and Tang, Hao and Wang, Weiyao and Guo, Michelle and Hardin, Thibaut and Li, Xiang and others},
  booktitle={Proceedings of the IEEE/CVF Conference on Computer Vision and Pattern Recognition},
  pages={7220--7232},
  year={2026}
}

@inproceedings{b24,
  title     = {SAM 3: Segment Anything with Concepts},
  author    = {Carion, Nicolas and Gustafson, Laura and Hu, Yuan-Ting and Debnath, Shoubhik and Hu, Ronghang and Suris, Didac and Ryali, Chaitanya and Alwala, Kalyan Vasudev and Khedr, Haitham and Huang, Andrew and Lei, Jie and Ma, Tengyu and Guo, Baishan and Kalla, Arpit and Marks, Markus and Greer, Joseph and Wang, Meng and Sun, Peize and R{\"a}dle, Roman and Afouras, Triantafyllos and Mavroudi, Effrosyni and Xu, Katherine and Wu, Tsung-Han and Zhou, Yu and Momeni, Liliane and Hazra, Rishi and Ding, Shuangrui and Vaze, Sagar and Porcher, Francois and Li, Feng and Li, Siyuan and Kamath, Aishwarya and Cheng, Ho Kei and Doll{\'a}r, Piotr and Ravi, Nikhila and Saenko, Kate and Zhang, Pengchuan and Feichtenhofer, Christoph},
  booktitle = {International Conference on Learning Representations (ICLR)},
  year      = {2026},
  url       = {https://openreview.net/forum?id=r35clVtGzw}
}

@inproceedings{PCN,
  author    = {Wentao Yuan and Tejas Khot and David Held and Christoph Mertz and Martial Hebert},
  title     = {PCN: Point Completion Network},
  booktitle = {Proceedings of the International Conference on 3D Vision (3DV)},
  year      = {2018},
  pages     = {728--737},
  doi       = {10.1109/3DV.2018.00088}
}

@inproceedings{GRNet,
  author    = {Haozhe Xie and Hongxun Yao and Shangchen Zhou and Jiageng Mao and Shengping Zhang and Wenxiu Sun},
  title     = {GRNet: Gridding Residual Network for Dense Point Cloud Completion},
  booktitle = {Proceedings of the European Conference on Computer Vision (ECCV)},
  year      = {2020},
  pages     = {365--381}
}

@inproceedings{GSPN,
  author    = {Li Yi and Wang Zhao and He Wang and Minhyuk Sung and Leonidas J. Guibas},
  title     = {GSPN: Generative Shape Proposal Network for 3D Instance Segmentation in Point Cloud},
  booktitle = {Proceedings of the IEEE/CVF Conference on Computer Vision and Pattern Recognition (CVPR)},
  year      = {2019},
  pages     = {3947--3956}
}

@inproceedings{BoNet,
  author    = {Bo Yang and Jie Wang and Ronald Clark and Qingyong Hu and Dylan Zeng and Liwen Wang and Andrew Markham and Niki Trigoni},
  title     = {Learning Object Bounding Boxes for 3D Instance Segmentation on Point Clouds},
  booktitle = {Advances in Neural Information Processing Systems (NeurIPS)},
  volume    = {32},
  year      = {2019}
}

@inproceedings{OccuSeg,
  author    = {Lei Han and Tian Zheng and Lan Xu and Lu Fang},
  title     = {OccuSeg: Occupancy-Aware 3D Instance Segmentation},
  booktitle = {Proceedings of the IEEE/CVF Conference on Computer Vision and Pattern Recognition (CVPR)},
  year      = {2020},
  pages     = {2938--2947}
}

@inproceedings{HAIS,
  author    = {Shaoyu Chen and Jiemin Fang and Qian Zhang and Wenyu Liu and Xinggang Wang},
  title     = {Hierarchical Aggregation for 3D Instance Segmentation},
  booktitle = {Proceedings of the IEEE/CVF International Conference on Computer Vision (ICCV)},
  year      = {2021},
  pages     = {15467--15476}
}

@inproceedings{Mask3D,
  author    = {Jonas Schult and Francis Engelmann and Alexander Hermans and Or Litany and Siyu Tang and Bastian Leibe},
  title     = {Mask3D: Mask Transformer for 3D Semantic Instance Segmentation},
  booktitle = {Proceedings of the IEEE International Conference on Robotics and Automation (ICRA)},
  year      = {2023},
  pages     = {8216--8223},
  doi       = {10.1109/ICRA48891.2023.10160590}
}


\end{document}